\documentclass[runningheads]{llncs}

\usepackage{latexsym}
\usepackage[T1]{fontenc}
\usepackage[utf8]{inputenc}
\usepackage{newunicodechar}
\newunicodechar{Ŀ}{\L.}
\PassOptionsToPackage{babel=true,expansion=true,kerning=true,tracking=true,spacing=true,verbose=silent}{microtype}

\usepackage{microtype}  
\usepackage{inconsolata}
\usepackage{graphicx}
\usepackage[toc,page]{appendix}
\usepackage{hyperref}
\usepackage{url}
\usepackage{siunitx}
\usepackage[most]{tcolorbox}
\usepackage{custom}
 
\definecolor{mygold}{RGB}{215, 155, 0}
\definecolor{myblue}{RGB}{108, 142, 191}
\definecolor{mygreen}{RGB}{0, 153, 0}

\title{Extraction and Analysis of Multimodal Concepts in Vision Language Models through Sparse Autoencoders}
\titlerunning{Extraction and Analysis of Multimodal Concepts in VLMs through SAEs }

\begin{document}
\author{Sergio Lanza\orcidID{0009-0004-8777-7070} \and
Jae Hee Lee\orcidID{0000-0001-9840-780X}\\ 
Stefan Wermter\orcidID {0000-0003-1343-4775}}
\authorrunning{S. Lanza et al.}
%
\institute{Knowledge Technology, Department of Informatics, 
\\ University of Hamburg, Hamburg, Germany\\
\email{sergio.lanza@uni-hamburg.de,
jae.hee.lee@uni-hamburg.de,
stefan.wermter@uni-hamburg.de}\\
\url{https://www.inf.uni-hamburg.de/en/inst/ab/wtm.html} 
}

\maketitle
\begin{abstract}

	Vision Language Models (VLMs) have demonstrated impressive performance in tasks requiring joint understanding of images and text, such as image captioning and Visual Question Answering (VQA), but our understanding of their internal processes remains limited. Recently, Sparse Autoencoders (SAEs) have emerged as a promising tool to support the interpretation of concepts encoded in VLMs. However, most SAE-based approaches focus only on textual or visual concepts separately, ignoring multimodal concepts.
	This limitation hinders a comprehensive understanding of VLMs, since concepts that integrate both modalities can be misclassified.
	Moreover, previous visual approaches often produce low-quality visual concept descriptions that are vague or incomplete, limiting their usefulness for understanding model reasoning. We propose a framework based on SAEs to extract and analyze visual, textual, and multimodal concepts from VLMs. For each neuron, we propose a candidate human-interpretable concept and compute the alignment between the concept and the dataset samples using cosine similarity scores. Experiments on a VQA dataset (LLaVA-NeXT) demonstrate that our framework improves visual concept quality by up to 45\% compared to existing SAE-based methods, while maintaining high textual concept quality and enabling systematic identification of multimodal concepts. This work contributes new insights into the conceptual space of VLMs, providing a structured approach to distinguish between visual, textual, and multimodal concepts.
    The code is available at \url{https://github.com/PHDLanza/Multidata_SAE}.

\end{abstract}

\section{Introduction}

Vision Language Models (VLMs) have achieved remarkable results in several tasks from image captioning to Visual Question Answering (VQA), yet understanding their internal mechanisms remains a critical challenge~\cite{kaduri2024whatsimagedeepdivevision,pach_sparse_2025}.
Mechanistic Interpretability (MI) addresses this challenge by studying the internal computation responsible for a model's output~\cite{bereska_mechanistic_2024}. One prominent direction within MI is Concept Extraction, which aims to identify human-understandable concepts encoded in the model activations to improve model transparency and interpretability~\cite{lee2024neuralactivationsconceptssurvey}.
Sparse Autoencoders (SAEs), in particular, have received great attention within Concept Extraction methods for extracting interpretable features encoded in pre-trained models, especially Large Language Models
(LLMs)~\cite{bricken_trenton_towards_nodate,cunningham_sparse_2023,elhage_toy_2022}.
SAEs project the high-dimensional embeddings of the LLM into a sparse representation composed of human-understandable features, or concepts, making this approach effective and scalable for concept extraction \cite{bereska_mechanistic_2024}.

However, most SAE-based works focus exclusively on textual concepts, with limited explorations of visual concepts and, to the best of our knowledge, no systematic study of multimodal concepts, i.e., concepts represented jointly in the visual and textual inputs. This gap limits our understanding of VLMs.
Recent studies suggest how VLMs encode multimodal information, but rely disproportionately on one modality depending on the task \cite{sim-etal-2025-vlms}. Understanding which information is encoded by which modality sheds light on the reasoning ability of VLMs and how they align and process inputs across modalities.
For example, a neuron may encode a concept that is jointly grounded in the image (an object or region) and in the question-answer text (a property, relation, or action).

In this paper, we present a novel concept extraction framework based on SAEs that integrates visual, textual, and multimodal analysis.
For each neuron in the SAE, we select the most representative images and textual samples that most strongly trigger it, then we infer candidate human-interpretable concepts from each modality, and finally, we evaluate which concept best represents the neuron's activation (see Figure \ref{fig:framework}).
The extraction is performed by an external VLM, LLaVA 72B, following prior works on SAE \cite{bricken_trenton_towards_nodate,cunningham_sparse_2023,gao_scaling_2024}. The evaluation, instead, is computed using CLIP \cite{radford_learning_2021} and ALIGN \cite{jia2021scalingvisualvisionlanguagerepresentation} similarities, which provides a shared space for text-image alignment~\cite{paulo_automatically_2024,radford_learning_2021}.

This work contributes new insights into the concepts encoded inside VLMs, providing a structured approach to distinguish between visual, textual, and multimodal concepts. Our novel contribution is divided into two main components:

\begin{itemize}
	\item \textbf{Integrated multimodal concept extraction:} We extend SAE analysis to extract visual, textual, and multimodal concepts in a unified framework, maintaining high concept quality across the three modalities.
	\item \textbf{Improved visual concept quality:}
	We introduce a new visual concept extraction technique that yields more precise and semantically grounded visual concepts. Our results on a VQA dataset, LLaVA-NeXT, show an improvement of up to 45\% in cosine similarity alignment between concept descriptions and their images compared to the existing SAE method.
\end{itemize}

\section{Background}
\label{sec:background}

Sparse Autoencoders (SAEs)~\cite{cunningham_sparse_2023} have been used to discover concepts encoded in the pretrained models. Specifically, it is an autoencoder that consists of a single high-dimensional hidden-layer (from 2$\times$ to over 100$\times$ the activation dimension) trained to reconstruct the activation from the model of interest. During the training phase, SAEs incentivize sparsity in the hidden layer, i.e., they minimize the number of activated neurons inside the layer.
Models that enhance sparsity help encode neurons representing a single, distinct concept called monosemantic neurons, which faithfully represent the backbone computation~\cite{elhage_toy_2022}.

As defined by~Pach et al.~\cite{pach_sparse_2025}, an SAE consists of two linear layers:
\begin{align*}
	W_{\text{enc}} & \in \mathbb{R}^{d \times \omega}, \quad W_{\text{dec}} \in \mathbb{R}^{\omega \times d},
\end{align*}
as encoder and decoder, respectively. The \(\omega\) term represents the dimension of the SAE hidden layer defined as \(\omega := d \cdot \varepsilon,\) where \( \varepsilon \) is the expansion factor and \( d\) the embedding dimension of the hosted model.
Given an input $x \in \mathbb{R}^{d}$, the SAE computes the latent activation $z  \in \mathbb{R}^{\omega}$  and reconstruction $\hat{x} \in \mathbb{R}^{d}$ as:
\begin{align*}
	z       & = \mathrm{TopK}\!\left(\mathrm{ReLU}\!\left(W_{\text{enc}}^\top(x - b) \right)\right) , \\
	\hat{x} & = W_{\text{dec}}^\top z + b,
\end{align*}
where $\mathrm{TopK}$ is the activation function that retains the highest K (a hyperparameter) values in the latent neurons while zeroing out all the others, enforcing sparsity in the latent representation, and \textit{b} is the bias term.
The SAE is trained to minimize the reconstruction loss:
\begin{align*}
	\mathcal{L}(x) & = \lVert x - \hat{x} \rVert_2
\end{align*}
encouraging faithful reconstruction of the original embedding based on the SAE neurons.
Since the hidden layer of an SAE can comprise more than 100k neurons, it is infeasible to manually associate each neuron with a concept based on the triggered inputs. The most commonly used approach involves collecting dataset samples (texts or images) that activate the same neuron and then querying an LLM, denoted as explainer, to infer a common concept
~\cite{bricken_trenton_towards_nodate,cunningham_sparse_2023,gao_scaling_2024}.

\section{Related Work}
\label{sec:related_work}

\subsection{Sparse AutoEncoders for VLMs}

The success of SAEs in the Language context has recently affected the concept extraction research for VLMs. Particularly, the approach of  Zhang et al.~\cite{zhang_large_2024} automatically detects and extracts concepts from individual SAE neurons injected into a VLM.
For each neuron in the latent space of the SAE, the top 5 images with the highest average activations are forwarded to the explainer, LLaVA 72B,~\cite{liu2024llavanext} that performs a zero-shot concept extraction task.
To ensure that the explainer focuses solely on the most interesting regions, they apply a masking process on the image patches: only the patches that activate the selected neuron are shown to LLaVA-72B, while all others are masked in black.
A related line of research studied the layers in the CLIP embedding \cite{lim_sparse_2025,pach_sparse_2025} using a similar masking technique and exploiting CLIP's semantic labelling capabilities to assess concepts within sub-patches of the image.
However, all these works do not consider textual concepts or analyze them superficially, limiting the insight that users can obtain, although all of them suggest the possible presence of these concepts in the VLM~\cite{lim_sparse_2025,pach_sparse_2025,zhang_large_2024}.
Moreover, the masking process removes too much contextual information, limiting the interpretability and reducing the accuracy of concepts extracted.
Our work addresses these limitations by jointly extracting and evaluating visual, textual, and multimodal concepts within a single SAE-based framework and improving the extraction process with additional textual information (see Section~\ref{sec:zero-shot}).

\section{Methodology}

\label{sec:methodology}
\subsection{Multimodal Concept Extraction \label{sec:zero-shot}}

\begin{figure}[!htbp]
	\centering
	\includegraphics[width=0.8\textwidth]{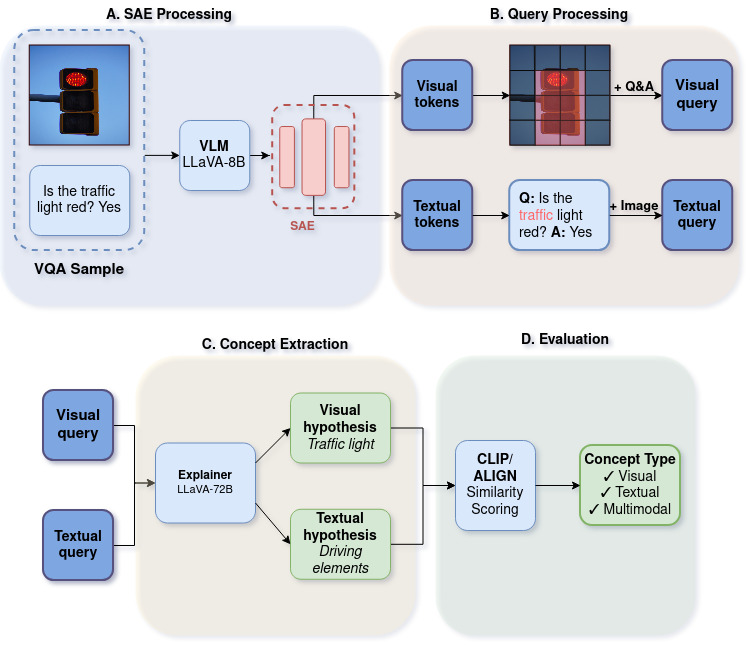}
	\caption{Overview of our multimodal concept extraction framework. \textbf{(A)} VQA sample is processed through a VLM with an integrated SAE that projects activations into a sparse hidden layer, producing visual and textual tokens. \textbf{(B)} For a target SAE neuron, we select the visual and textual tokens that strongly activate it and construct modality-specific inputs: non-activating visual tokens (or patches) are black masked, while activating textual tokens are highlighted. From these inputs, we construct two queries, each augmented with complementary information (the question–answer pair for visual extraction and the unmodified image for textual extraction). \textbf{(C)} An external VLM (the explainer) extracts concept hypotheses from the queries. \textbf{(D)} CLIP/ALIGN evaluates hypothesis data alignment to classify each neuron as visual, textual, or multimodal.}
	\label{fig:framework}
\end{figure}

Inspired by previous work \cite{lim_sparse_2025,zhang_large_2024}, we design our novel concept hypothesis extractor leveraging an external VLM as an explainer.
All experiments are conducted on a VQA dataset; VQA is a task where a model answers a question about an image and is evaluated against a ground-truth answer.
First, we feed a set of VQA samples into a VLM augmented with a pretrained SAE to record which sample strongly activates the studied SAE neurons.
For each VQA sample \(j\), which comprises an image and a textual question--answer pair, we record two separate SAE activation tensors corresponding to visual and textual tokens defined as:

\begin{align*}
	V^{(j)} = [v^{(j)}_1, \ldots, v^{(j)}_P]\in \mathbb{R}^{P \times \omega} \;, \qquad T^{(j)} = [t^{(j)}_1, \ldots, t^{(j)}_{N_j}] \in \mathbb{R}^{N_j \times \omega} \;
\end{align*}
where \(v^{(j)}_{i}\) and \(t^{(j)}_{i}\) are the activation vectors for the $i$-th visual patch or textual token respectively, \(P\) represents the number of visual patches per image (24 \(\times\) 24), \(N_{j}\) corresponds to the number of textual tokens encoded in the $j$-th sample and \(\omega\) denotes the hidden dimension of SAE. From these matrices, we compute the average activations normalized for patches, $\bar{V}^{(j)} \in \mathbb{R}^{\omega}$,  or tokens, $\bar{T}^{(j)} \in \mathbb{R}^{\omega}$, defined as:
\begin{align*}
	\bar{V}^{(j)} = \frac{1}{P} \sum_{i=1}^{P} v^{(j)}_i, \qquad \bar{T}^{(j)} = \frac{1}{N_j} \sum_{i=1}^{N_j} t^{(j)}_i
\end{align*}
For each SAE neuron \(k\), we then select the top-5 samples that maximize \(\bar{V}^{(j)}_k\) (visual) and \(\bar{T}^{(j)}_k\) (textual), and query the explainer to extract the hypothesis. As illustrated in Figure~\ref{fig:framework}, in the visual extraction process, we apply a black mask to the patches that do not activate the targeted neuron,  following prior studies~\cite{pach_sparse_2025,zhang_large_2024}, while including the corresponding VQA question and answer alongside the image input. We introduce this element following evidence that additional related text improves visual understanding accuracy in VLMs~\cite{ma2024doesvlmclassificationbenefit,zang2025pretrainedvisionlanguagemodelslearn}. For textual extraction, we apply a similar procedure: we highlight the specific tokens that trigger the neuron by enclosing them in brackets and pass these sentences to the external VLM, including the corresponding VQA images. This approach mirrors the visual process and provides richer multimodal grounding in the textual extraction.

\subsection{Metrics}
\label{metric}

After the extraction of the concept hypotheses (see Section~\ref{sec:zero-shot}), each SAE neuron yields up to two hypotheses: visual, textual, or no hypothesis if the available data are insufficient to produce a shared concept.

To quantitatively evaluate the hypotheses, we require a precise and modality-independent measure.
We therefore adopt CLIP ViT-B/32~\cite{radford_learning_2021} and ALIGN~\cite{jia2021scalingvisualvisionlanguagerepresentation} as our evaluators, two vision-language models trained to align images and text in a shared embedding space, which have been used in prior interpretability works~\cite{zhang_large_2024,pach_sparse_2025}.
We utilize CLIP and ALIGN in two evaluation settings.
\paragraph{Hypothesis–data alignment}
In the first setting, for each hypothesis and each evaluator, we input to the model the same data used in the extraction, specifically blurred image patches for visual hypotheses and token sequences for textual hypotheses, and compute the similarity score for all the hypotheses. In preliminary experiments, we observed that black masking degraded the CLIP and ALIGN reliability; therefore, we blurred the inactive patches, leading to more stable and interpretable outcomes.
We obtain a similarity matrix of scores for each modality comparing the sets of data used in the extraction process and all the hypotheses, and we rank the highest score for each row. This procedure assesses the alignment between the hypothesis and the data from which it was generated.
To account for semantic similarity among hypotheses, we consider a hypothesis validated if its score on the data that generated it ranks within the top \(n\) predictions.
In our experiments, we use $n \in \{1, 2, \ldots, 5\}$.

\paragraph{Multimodal neuron assignment}
For neurons with both a visual hypothesis \(h_{\text{vis}}\) and a textual hypothesis \(h_{\text{text}}\), we assign a modality label based on semantic consistency between the two hypotheses.
We encode \(h_{\text{vis}}\) and \(h_{\text{text}}\) using the text encoders of CLIP and ALIGN and compute cosine similarity.
We label a neuron as \textit{multimodal} if both similarities exceed a threshold \(\tau\) (\(\tau=0.8\) in our experiments); otherwise, we assign it to the modality whose hypothesis yields the higher hypothesis--data alignment score.
In Section~\ref{app:tao}, we report a sensitivity analysis over other \(\tau\) values.

\paragraph{Category-based evaluation}

For the second setting, we follow the evaluation procedure of Zhang et al.~\cite{zhang_large_2024} to ensure a direct comparison with our baseline. Each visual hypothesis is mapped to a specific visual category, and then we compute the CLIP and ALIGN scores between the samples and the assigned category label.
An external VLM performs the category mapping, and the six categories follow the taxonomy introduced in \cite{bau_network_2017}.
While the first setting offers a more fine-grained evaluation and explicitly accounts for textual and multimodal concepts, we also adopt the second to enable a fair comparison with our baseline (see Section \ref{sec:experiment}), which focuses solely on visual concepts.

\paragraph{Additional textual evaluation}

Finally, we assess the quality of textual extraction to verify that textual hypotheses are not biased by the inclusion of visual information during the extraction procedure. Following prior works described in Section \ref{sec:background}, we query an external LLM explainer with randomly sampled textual sentences from LLaVA-NeXT together with the analyzed hypothesis. The explainer predicts whether the hypothesis is effectively represented in the sentences, and then we compute the accuracy and the precision against the ground-truth annotations.
We utilize the evaluation protocol of Paulo et al.~\cite{paulo_automatically_2024}, specifically, we adopt the detection score, which evaluates hypothesis presence at the sentence level, and the fuzzing score, focused on hypothesis presence at the token level.

\section{Experiments}
\label{sec:experiment}

In this section, we present the details of the quality and modality distribution of extracted concepts. We compared our results to Zhang et al.~\cite{zhang_large_2024}, referring to it as our baseline. To be consistent with the baseline, our analysis focuses only on the first 5000 neurons of the SAE.

\subsection{Models}
For consistency with prior work~\cite{zhang_large_2024}, we adopt the same VLMs and SAE.
The SAE is trained on frozen embeddings extracted from the LLaVA-NeXT-8B-HF model \cite{liu2024llavanext} with Top-k activation (\(k=256\)) and a latent space \(\omega=2^{17}\) neurons. Based on prior work, we select the SAE trained on the 24$^{\text{th}}$ transformer layer, which empirically yields the highest visual interpretability.
For concept hypothesis extraction as an explainer, we adopt LLaVA-72B and select the five best samples to remain consistent with our baseline.
For the textual analysis only, we utilize Llama-3.1-70B-Instruct~\cite{grattafiori2024llama3herdmodels}, an LLM already adopted in another textual SAE analysis work~\cite{paulo_automatically_2024} to avoid undesired visual bias that the VLM could bring.
All experiments are run on two NVIDIA H100 GPUs: a complete run of the framework, from activation extraction to concept evaluation, requires approximately 30 hours of computation.

\begin{table}[h]
	\centering
	\caption{Comparison of correct visual concept extraction rates between our framework (\textbf{Ours}) and the \textbf{baseline}~\cite{zhang_large_2024} using CLIP and ALIGN scores. Our method outperforms the baseline at all thresholds by an average margin of 41\% (CLIP) and 45\% (ALIGN).}
	\label{table:CLIP_ALIG}
	\setlength{\tabcolsep}{4pt}
	\begin{tabular}{lcccc}
		\toprule
		\textbf{Thresh.}      &
		\textbf{Base (CLIP)}  &
		\textbf{Ours (CLIP)}  &
		\textbf{Base (ALIGN)} &
		\textbf{Ours (ALIGN)}                                                                             \\

		\cmidrule(lr){2-2} \cmidrule(lr){3-3} \cmidrule(lr){4-4} \cmidrule(lr){5-5}
		                      & Avg ($\uparrow$) & Avg ($\uparrow$) & Avg ($\uparrow$) & Avg ($\uparrow$) \\
		\midrule

		1                     & 0.15             & \textbf{0.45}    & 0.13             & \textbf{0.47}    \\
		2                     & 0.22             & \textbf{0.61}    & 0.20             & \textbf{0.65}    \\
		3                     & 0.26             & \textbf{0.69}    & 0.24             & \textbf{0.69}    \\
		4                     & 0.29             & \textbf{0.75}    & 0.26             & \textbf{0.74}    \\
		5                     & 0.31             & \textbf{0.79}    & 0.27             & \textbf{0.80}    \\
		\midrule
		\textbf{Avg}          & 0.246            & \textbf{0.658}   & 0.220            & \textbf{0.670}   \\
		\bottomrule
	\end{tabular}
\end{table}

\subsection{Datasets}
\label{sec:dataset}

To remain consistent with our visual baseline, our main experiment uses the LLaVA-NeXT dataset \cite{liu2024llavanext}, which follows the VQA format: each sample comprises an image, a corresponding question, and an answer.
We also replicate our experiments on a subset of the COCO dataset \cite{dong_huk_vqa} to evaluate the generalizability of our approach with a different set of data (see Appendix \ref{app:COCO_dataset}). Due to computational constraints and to remain consistent with our baseline, we use a subset of it, specifically the first 15\% of LLaVA-NeXT ~\cite{zhang_large_2024}.
All prompts used are reported in the Appendix \ref{app:prompts}.


\subsection{Multimodal Extraction}
\label{sec:multi}

\begin{figure}
	\centering

	\includegraphics[width=0.8\linewidth]{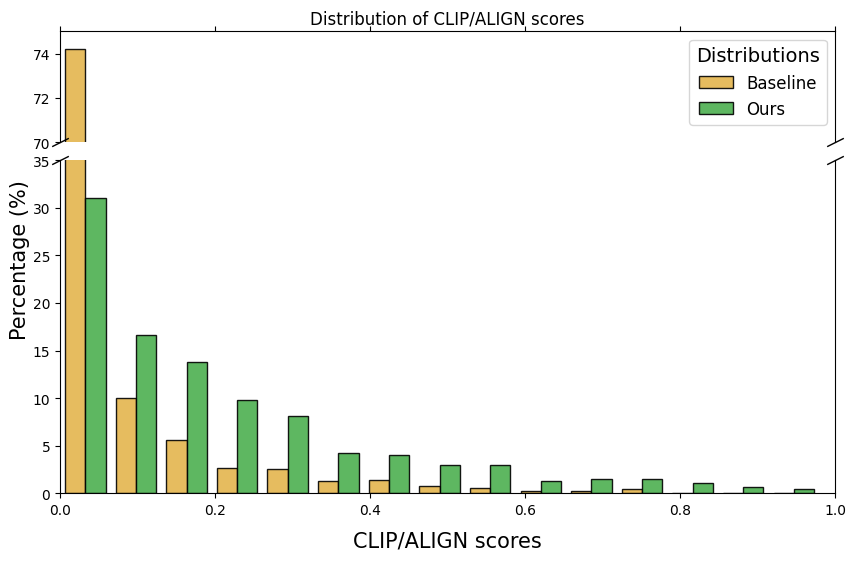}
	\caption{Distribution of the CLIP and ALIGN scores for the proposed model (Ours) compared to the baseline. For each sample, we compute the average between the CLIP and ALIGN scores. The histogram illustrates the percentage of samples across similarity score intervals, highlighting a shift toward higher average similarity values in the proposed method relative to the baseline distribution.}
	\label{fig:histogram}
\end{figure}

In Table \ref{tab:results}, we show the number of concept hypotheses extracted by our framework and the baseline (see Appendix \ref{app:concepts} for some qualitative examples).
Although our framework extracts fewer visual concepts compared to the baseline, we argue that the concepts identified by our approach are more semantically representative of the images utilized to extract them.
To validate this claim, we compute the direct concept alignment scores following the procedures described in Section~\ref{metric}. As illustrated in Table~\ref{table:CLIP_ALIG}, our approach consistently outperforms the baseline with higher accuracy in all thresholds and in both metrics, indicating a stronger alignment between images and the identified visual and multimodal concepts.
To confirm this, Figure~\ref{fig:histogram} shows that our approach identifies fewer concepts with low CLIP scores compared to the baseline.
Instead, in Table~\ref{tab:cos_table}, our framework achieves comparable or slightly superior results in the category-concept alignment score.
Finally, in Table~\ref{table:auto} we assessed the quality of textual concepts using the Interpretability score~\cite{paulo_automatically_2024}. We compare our approach with other similar SAEs using only the textual approach, obtaining similar results.

\begin{table}[!htbp]
	\caption{Average accuracy of correct visual concept categorization through CLIP and ALIGN similarities with \cite{bau_network_2017} labels.}
	\centering
	\small 
	\setlength{\tabcolsep}{3pt} 
	\label{tab:cos_table}
	\resizebox{\columnwidth}{!}{
	\begin{tabular}{lcccccccc}
		\toprule
		\textbf{Label} & \multicolumn{2}{c}{\textbf{Baseline (CLIP)}} & \multicolumn{2}{c}{\textbf{Ours (CLIP)}} & \multicolumn{2}{c}{\textbf{Baseline (ALIGN)}} & \multicolumn{2}{c}{\textbf{Ours (ALIGN)}}                                                                                 \\
		\cmidrule(lr){2-3} \cmidrule(lr){4-5} \cmidrule(lr){6-7} \cmidrule(lr){8-9}
		               & Avg ($\uparrow$)                             & Var ($\downarrow$)                       & Avg ($\uparrow$)                              & Var ($\downarrow$)                        & Avg ($\uparrow$) & Var ($\downarrow$) & Avg ($\uparrow$) & Var ($\downarrow$) \\
		\midrule
		scene          & 18.84                                        & $9.4\times10^{-3}$                        & \textbf{26.58}                                & $1.3\times100^{-2}$                         & 16.84            & $1.5\times10^{-6}$  & \textbf{16.94}   & $1.4\times10^{-6}$  \\
		object         & \textbf{40.65}                               & $2.3\times10^{-2}$                        & 39.02                                         & $2.2\times10^{-2}$                         & 16.69            & $1.4\times10^{-6}$  & \textbf{16.73}   & $1.4\times10^{-6}$  \\
		part           & \textbf{23.24}                               & $8.7\times10^{-3}$                        & 20.93                                         & $5.0\times10^{-3}$                         & \textbf{16.69}   & $1.0\times10^{-6}$  & 16.66            & $1.8\times10^{-6}$  \\
		material       & \textbf{5.36}                                & $1.0\times10^{-3}$                        & 4.98                                          & $6.0\times10^{-4}$                         & \textbf{16.59}   & $1.0\times10^{-6}$  & 16.57            & $1.8\times10^{-6}$  \\
		texture        & 4.69                                         & $3.0\times10^{-3}$                        & \textbf{9.61}                                 & $8.4\times10^{-3}$                         & 16.69            & $1.9\times10^{-6}$  & \textbf{16.86}   & $1.8\times10^{-6}$  \\
		color          & \textbf{32.27}                               & $2.8\times10^{-2}$                        & 25.76                                         & $2.1\times10^{-2}$                         & \textbf{16.82}   & $2.1\times10^{-6}$  & 16.69            & $2.0\times10^{-6}$  \\
		\midrule
		\textbf{Avg}   & 20.84                                        & $1.2\times10^{-2}$                        & \textbf{21.98}                                & $1.1\times10^{-2}$                         & 16.72            & $1.4\times10^{-6}$  & \textbf{16.74}   & $1.6\times10^{-6}$  \\
		\bottomrule
	\end{tabular}
	}
\end{table}

\begin{table*}[ht!]
	\centering
	\small
	\begin{minipage}[t]{0.32\textwidth}
		\centering
		\caption{Number of concept hypotheses from our framework and the baseline~\cite{zhang_large_2024}.}
		\begin{tabular}{l c c}
			\toprule
			\textbf{Modality} & \textbf{Zhang} & \textbf{Ours} \\
			\midrule
			Textual           & --             & 2291          \\
			Visual            & 1391           & 470           \\
			Multimodal        & --             & 132           \\
			\bottomrule
		\end{tabular}
		\label{tab:results}
	\end{minipage}
	\hfill
	\begin{minipage}[t]{0.32\textwidth}
		\centering
		\caption{Interpretability scores for textual concepts~\cite{paulo_automatically_2024}. Average of 5~runs.}
		\begin{tabular}{l c c}
			\toprule
			\textbf{Models} & \textbf{Det.} & \textbf{Fuzz.} \\
			\midrule
			Gemma 2 9b      & 0.76          & 0.74           \\
			Llama 3.1 8b    & 0.81          & 0.83           \\
			LLaVA (Ours)    & 0.78          & 0.80           \\
			\bottomrule
		\end{tabular}
		\label{table:auto}
	\end{minipage}
	\hfill
	\begin{minipage}[t]{0.32\textwidth}
		\centering
		\caption{Multimodal concept counts at different thresholds (249 SAE neurons).}
		\begin{tabular}{l c}
			\toprule
			\textbf{Threshold} & \textbf{Count} \\
			\midrule
			0.70               & 223            \\
			0.80               & 132            \\
			0.90               & 23             \\
			\bottomrule
		\end{tabular}
		\label{tab:threshold_analysis}
	\end{minipage}
\end{table*}

\subsection{Analysis for multimodal threshold}
\label{app:tao}

In this section, we analyze the effect of different values of the threshold \(\tau\). Table~\ref{tab:threshold_analysis} reports the number of neurons whose visual and textual hypotheses exceed each threshold for both CLIP and ALIGN.
As expected, increasing \(\tau\) results in a stricter criterion and, as a consequence, a smaller number of multimodal concepts. Conversely, decreasing \(\tau\) increases the number of multimodal concepts.
Based on this analysis, we select an intermediate threshold of \(\tau = 0.8\) which balances between coverage and semantic consistency.

\subsection{Qualitative Intervention}

\begin{figure}[!htbp]
	\centering
	\includegraphics[width=1\linewidth]{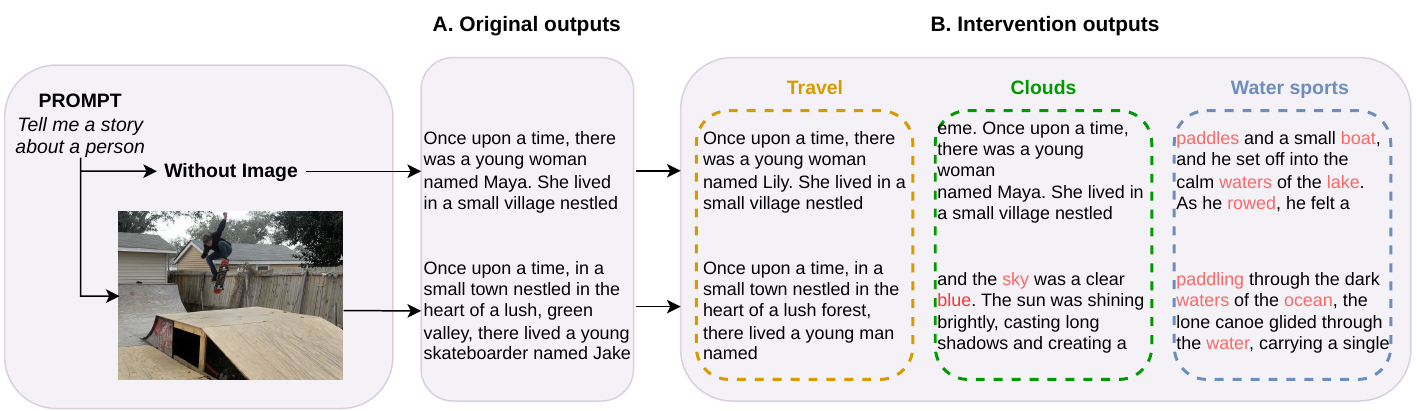}
	\caption{Qualitative example for intervention experiments using the prompt ``\textbf{Tell me a story about a person}''. \textbf{(A)} We generate two outputs, with and without an image prompt. \textbf{(B)} We show six intervention outputs, where three concepts are injected independently for each prompt. The concepts are selected from the three categories: \textcolor{mygold}{textual}, \textcolor{mygreen}{visual}, and \textcolor{myblue}{multimodal}. Words referring to the injected concept are highlighted in \textcolor{red}{red}.}
	\label{fig:intervention}
\end{figure}

In this experiment, we want to evaluate the impact of specific concepts on text generation qualitatively \cite{pach_sparse_2025}.
We tested the LLaVA-NeXT-8B-HF model, already integrated with our SAE, with the instruction ``\textbf{Tell me a story about a person}''. The prompt is provided in two versions: one with an associated image as additional input and one without, to observe how visual information affects the intervention.
During the new generation, we force the target neuron's activation to the fixed value \(\alpha=60\), nearly 20 times the average activation of an SAE neuron, following past results~\cite{pach_sparse_2025}.
We select three concepts, one from each modality, to analyze which modality has the biggest impact on text generation. As illustrated in Figure~\ref{fig:intervention}, the output modifications are more pronounced generally when the input includes an image, suggesting that visual features amplify the effect of concept-level intervention, and meanwhile multimodal concepts seem to affect the output production more than the other two categories.

\section{Discussion}
Our work primarily focuses on a single SAE and a single VLM as an explainer.
Future work will extend the generalizability of the proposed framework across multiple architectures.
Our concepts are generated from two independent prompts, designed to produce concise and interpretable descriptions rather than longer, free-form explanations. While this choice improves consistency and evaluation reliability, particularly because CLIP and ALIGN are more effective on short-medium length texts, it may restrict the diversity and semantic richness of the extracted concepts.
Finally, our findings reveal a clear correlation between SAE's concept alteration with the text output, but they do not establish a causal relationship.

\section{Conclusion}

In this paper, we demonstrate that an SAE encodes textual and visual multimodal concepts condensed into a shared latent space. Based on this property, we design a novel framework that performs dual-modality concept extraction, enriching both visual and textual hypotheses with multimodal information.
Our results exhibit a significant improvement in the quality of interpretable SAE neurons connected to visual concepts, achieving an average improvement of 41\% and 45\% for CLIP and ALIGN scores, maintaining good quality textual production, and providing a more comprehensive understanding of the internal reasoning of VLMs.
Finally, our intervention studies confirm the influence of multimodal and visual concepts on text generation. Overall, this work offers a new understanding of the concept space representation of VLMs by leveraging a new SAE-based approach. We hope this work encourages future research on scalable SAE integration, multimodal interpretability, and concept-guided control in multimodal reasoning systems.

\section{Acknowledgments}
The authors gratefully acknowledge funding from Horizon Europe under the MSCA grant agreements No 101072488 (TRAIL),No 101226624 (GREET), and No 101168792 (SWEET), as well as from the German Research Foundation (DFG), project number 551629603.

\bibliographystyle{splncs04}
\bibliography{custom}

\appendix
\label{sec:appendix}
\section{Analysis on COCO dataset}
\label{app:COCO_dataset}
In this section, we report additional results of our framework applied to a different VQA dataset based on COCO\cite{coco-VQA}, using the SAE injected into LLaVA-NeXT-8B-HF model. 
Although SAEs are generally not conditioned on the dataset used during inference, with this test, we verify whether our framework preserves this property for the first 1000 neurons.
\
\begin{table}[ht]
    \caption{COCO generalization results for the first 1000 neurons.}
    \centering
    \small
    \begin{tabular}{lccc}
        \toprule
        \textbf{Dataset} & \textbf{\#Text} & \textbf{\#Visual} & \textbf{\#Multimodal} \\
        \midrule
        LLaVA-NeXT (main) & 639 & 85 & 19 \\
        COCO (subset) & 537 & 57 & 9 \\
        \bottomrule
    \end{tabular}
    \label{tab:coco_results}
\end{table}

We also compare concept hypotheses across datasets using CLIP and ALIGN embeddings to estimate the overlap and stability of extracted concepts.
\begin{table}[ht]
    \caption{Average similarity scores between the two dataset}
    \centering
    \small
    \begin{tabular}{lccc}
        \toprule
        \textbf{Metric} & \textbf{LLaVA-NeXT} & \textbf{COCO}  \\
        \midrule
        CLIP & 639 & 85  \\
        ALIGN & 537 & 57 \\
        \bottomrule
    \end{tabular}
    \label{tab:clip_align_coco}
\end{table}

\section{Qualitative Examples}
\label{app:concepts}

\begin{figure}[htbp]
	\centering
	\includegraphics[width=.7\linewidth]{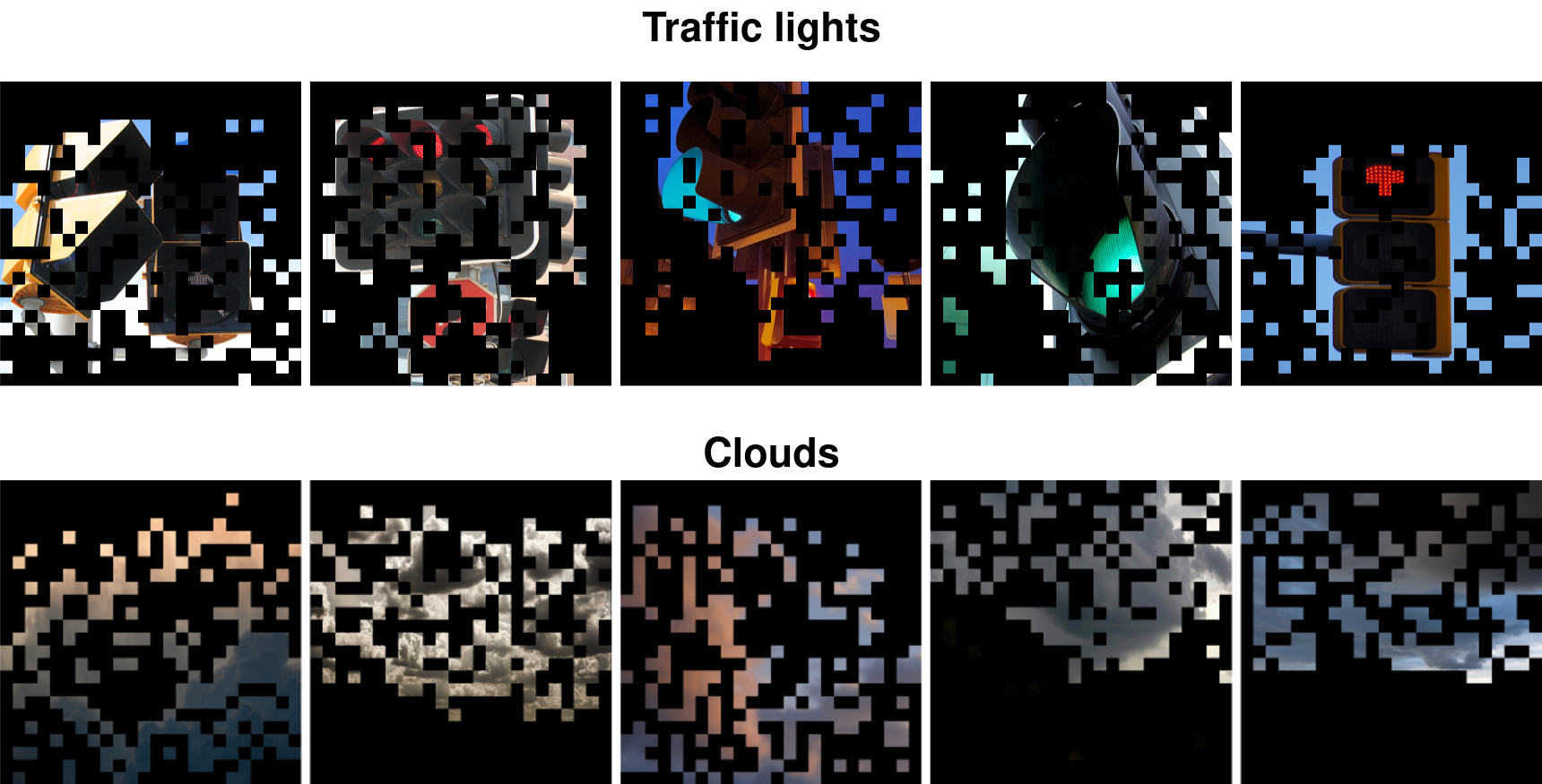}
	\caption{Examples of visual concepts.}
	\label{fig:visual_concepts}
\end{figure}

\begin{figure}[htbp]
	\centering
	\includegraphics[width=.7\linewidth]{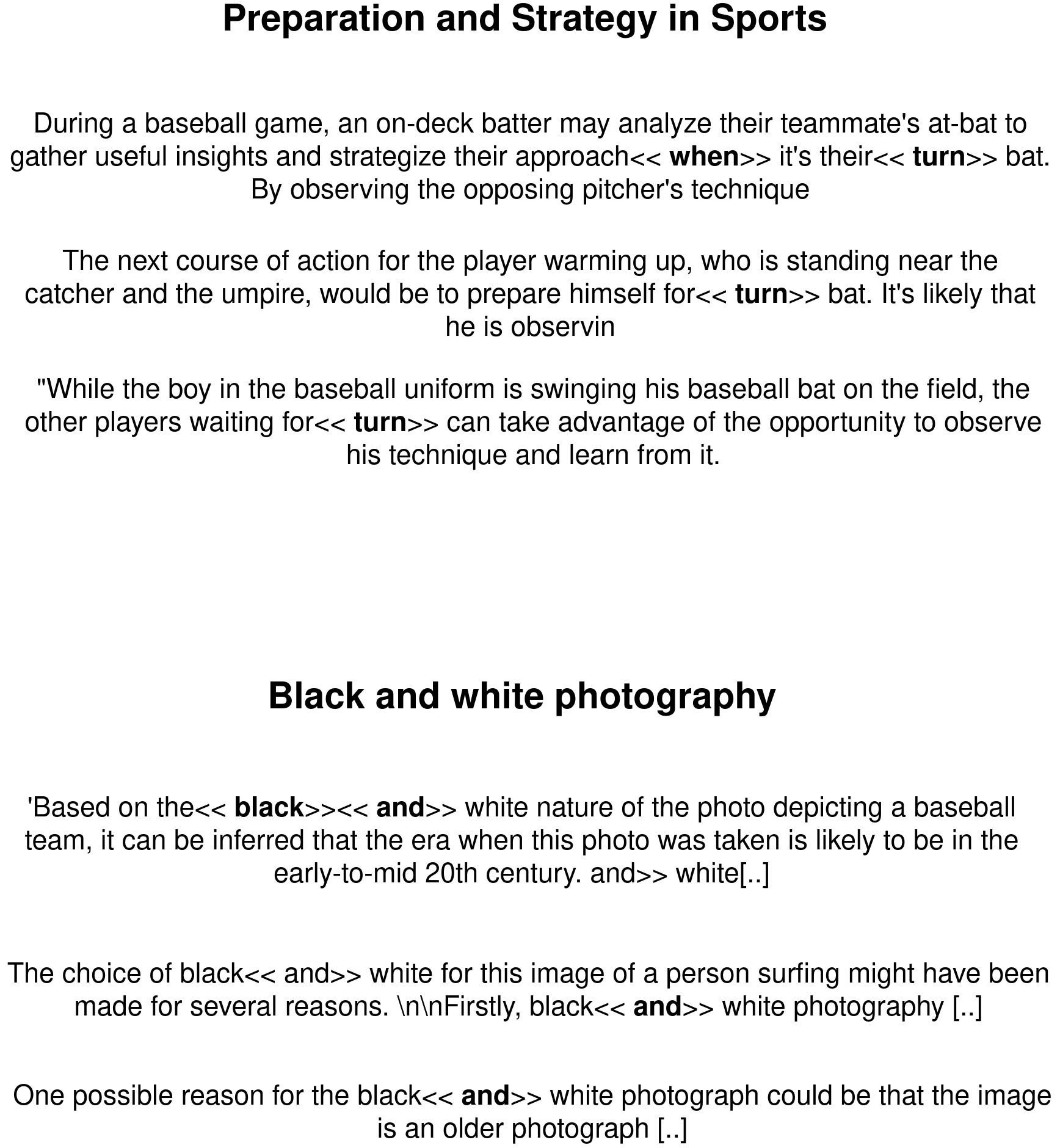}
	\caption{Examples of textual concepts.The textual tokens that trigger the SAE neuron are highlighted with <<>>.}
	\label{fig:textual_concepts}
\end{figure}

In this section, we present qualitative examples of concepts grouped by modality (Figures~\ref{fig:multimodal_concepts},~\ref{fig:visual_concepts}, and \ref{fig:textual_concepts}):
\begin{figure}[htbp]
	\centering
	\includegraphics[width=.7\linewidth]{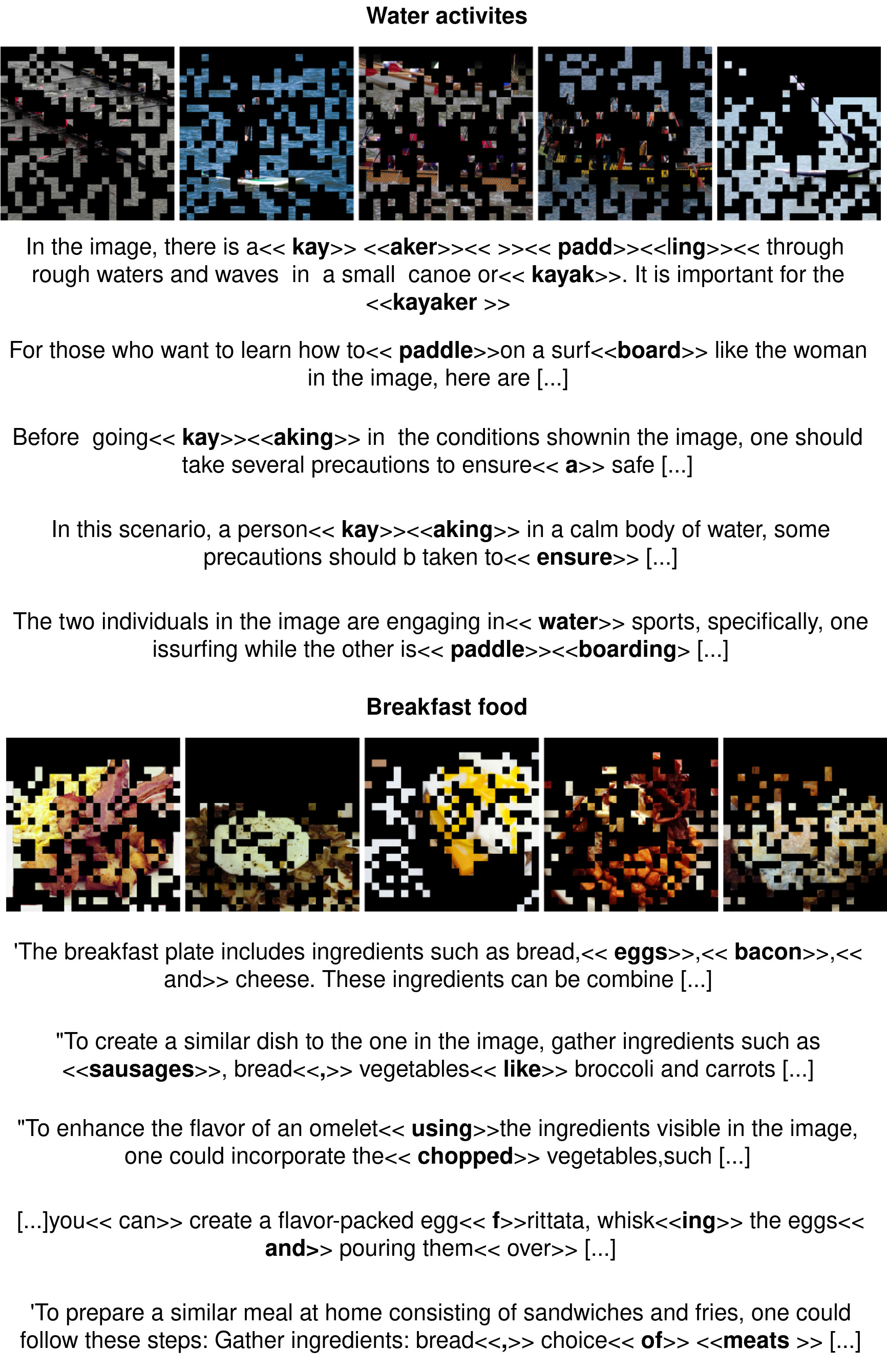}
	\caption{Examples of multimodal concepts. The textual tokens that trigger the SAE neuron are highlighted with <<>>.}
	\label{fig:multimodal_concepts}
\end{figure}

\section{Prompts}
\label{app:prompts}

Below are the prompts used to query the LLaVA-72B model (Figures~\ref{fig:llava_prompt_labelling},~\ref{fig:Llava_prompt_visual},~\ref{fig:Llava_prompt_textual}), inspired by or directly taken from~\cite{zhang_large_2024}. The first prompt maps each visual hypothesis to one of six predefined labels, and the next two prompts generate visual and textual hypotheses (see Section~\ref{sec:zero-shot}).

\begin{figure}[!htbp]
	\centering
	\small
	\begin{tcolorbox}[width=\linewidth, colframe=black, colback=gray!5, boxrule=0.5pt]
		[GUIDELINES]

		You are an AI assistant tasked with assigning a single label based on the given input text. Each input will contain a description of a visual feature, which you must categorize into one of the following classes:

		scene - Describes a scene or environment.

		object - Describes an object or entity.

		part - Describes a part or aspect of an object.

		material - Describes a material or substance that constitutes other objects.

		texture - Describes the texture of an object.

		color - Describes the color of an object.

		Please provide only the class label from the list [scene, object, part, material, texture, color] with no additional text.

		Only one label should be chosen. Make sure you only choose from the classes listed above and do not output any other classes.

		Categorize the following description: \{description\}

		ANSWER:
	\end{tcolorbox}

	\caption{Labelling Prompt (LLaVA 72B) }
	\label{fig:llava_prompt_labelling}
\end{figure}

\begin{figure}[!htbp]
	\centering
	\small
	\begin{tcolorbox}[width=\linewidth, colframe=black, colback=gray!5, boxrule=0.5pt]
		[REQUIREMENTS]

		Focus only on the highlighted region in each image. If no region is highlighted or if the highlighted region is minimal (e.g., a few bright spots), ignore the image.

		Identify common visual patterns, objects, or concepts in the activated regions. For example, note if highlighted areas show consistent structures, such as mesh patterns or similar objects.

			[GUIDELINES]

		1. Consider Text Context: While maintaining primary focus on the highlighted regions in images, you may marginally consider the associated text (questions and answers) to support or refine your visual observations.
		However, the final concept should be predominantly based on visual patterns.

		2. Concise Description Only: Provide a short, direct description of the common features within the highlighted regions. Avoid any interpretive language—simply state what you see, such as “mesh-like structures” or “actions related to joy or happiness”.

		3. Describe Only the Highlighted Regions: Generate captions solely based on the highlighted regions. If no meaningful pattern is visible, or if only a few scattered spots are highlighted, output: "Concept: `No visual concept`"
	\end{tcolorbox}

	\caption{Visual Generation Guidelines (LLaVA 72B) }
	\label{fig:Llava_prompt_visual}
\end{figure}


\begin{figure}[!htbp]
	\centering
	\small
	\begin{tcolorbox}[width=\linewidth, colframe=black, colback=gray!5, boxrule=0.5pt]
		[REQUIREMENTS]

		Focus only on the text content provided with each example. If the text is missing, irrelevant, or extremely minimal (e.g., a few unrelated words), ignore that example.

		Identify common themes, objects, or concepts mentioned across the text snippets. Pay special attention to any highlighted word in each text—this word should be treated as the most important cue for concept identification.

			[GUIDELINES]

		1. You will receive a series of text snippets, sometimes accompanied by images. Only use the text, and in particular the word between parentheses, to identify the shared concept. Images should not be considered in your analysis.

		These examples are derived from a Visual Question Answering dataset, so each text is in the form of a question or an answer.

		2. Concise Description Only: Provide a short, direct description of the common concept emerging from the texts. Avoid speculation or abstract interpretation—simply state what is explicitly or implicitly repeated, especially in relation to the highlighted words (e.g., “vehicles,” “cooking actions,” “types of animals”).

		Use the image only for reference if absolutely necessary; the main analysis must be text-driven, with words in parentheses as priority.

		3. If no clear concept emerges from the texts (e.g., if they are too diverse or vague), write: No textual concept

			[OUTPUT EXAMPLES]

		Concept: "A tennis match"

		Concept: "Descriptions of birds"

		Concept: "No textual concept"

		Remember, Write always only one Concept for the entire set of inputs
	\end{tcolorbox}

	\caption{A4: Textual Generation Guidelines (LLaVA 72B)}
	\label{fig:Llava_prompt_textual}

\end{figure}

The last two prompts are used to query Llama 3.1 70B, which evaluates textual concepts (Figures~\ref{fig:Llama_prompt_det} and~\ref{fig:Llama_prompt_fuz}) based on~\cite{paulo_automatically_2024}.

\begin{figure}[!htbp]
	\centering
	\small
	\begin{tcolorbox}[width=\textwidth, colframe=black, colback=gray!5, boxrule=0.5pt]
		"""You are an intelligent and meticulous linguistics researcher.

		You will be given a certain latent of text, such as "male pronouns" or "text with negative sentiment".

		You will then be given several text examples. Your task is to determine which examples possess the latent.

		For each example in turn, return 1 if the sentence is correctly labeled or 0 if the tokens are mislabeled. You must return your response in a valid Python list. Do not return anything else besides a Python list.

		Latent explanation: Words related to American football positions, specifically the tight end position.

		Test examples:

		Example 0:<|endoftext|>Getty ImagesĊĊPatriots tight end Rob Gronkowski had his bossâĢĻ
		Example 1: names of months used in The Lord of the Rings:ĊĊâĢľâĢ¦the
		Example 2: Media Day 2015ĊĊLSU defensive end Isaiah Washington (94) speaks to the
		Example 3: shown, is generally not eligible for ads. For example, videos about recent tragedies,
		Example 4: line, with the left side âĢĶ namely tackle Byron Bell at tackle and guard Amini

		"[1,0,0,0,1]"

		Latent explanation: The word "guys" in the phrase "you guys".

		Test examples:

		Example 0: enact an individual health insurance mandate?âĢĿ, Pelosi's response was to dismiss both
		Example 1: birth control access<|endoftext|> but I assure you women in Kentucky aren't laughing as they struggle
		Example 2: du Soleil Fall Protection Program with construction requirements that do not apply to theater settings because
		Example 3:Ċ<|endoftext|> distasteful. Amidst the slime lurk bits of Schadenfre
		Example 4: the<|endoftext|>ľI want to remind you all that 10 days ago (director Massimil

		"[0,0,0,0,0]"

		Latent explanation: "of" before words that start with a capital letter.

		Test examples:

		Example 0: climate, TomblinâĢĻs Chief of Staff Charlie Lorensen said.Ċ
		Example 1: no wonderworking relics, no true Body and Blood of Christ, no true Baptism
		Example 2:ĊĊDeborah Sathe, Head of Talent Development and Production at Film London,
		Example 3:ĊĊIt has been devised by Director of Public Prosecutions (DPP)
		Example 4: and fair investigation not even include the Director of Athletics? A· Finally, we believe the

		"[1,1,1,1,1]"
		"""

	\end{tcolorbox}

	\caption{ Detection score prompt (Llama 3.1 70B)}
	\label{fig:Llama_prompt_det}

\end{figure}

\begin{figure}[!htbp]
	\centering
	\small
	\begin{tcolorbox}[width=\textwidth, colframe=black, colback=gray!5, boxrule=0.5pt]

		"""
		You are an intelligent and meticulous linguistics researcher.

		You will be given a certain latent of text, such as "male pronouns" or "text with negative sentiment". You will be given a few examples of text that contain this latent. Portions of the sentence which strongly represent this latent are between tokens << and >>.

		Some examples might be mislabeled. Your task is to determine if every single token within << and >> is correctly labeled. Consider that all provided examples could be correct, none of the examples could be correct, or a mix. An example is only correct if every marked token is representative of the latent

		For each example in turn, return 1 if the sentence is correctly labeled or 0 if the tokens are mislabeled. You must return your response in a valid Python list. Do not return anything else besides a Python list.
		Latent explanation: Words related to American football positions, specifically the tight end position.

		Test examples:

		Example 0:<|endoftext|>Getty ImagesĊĊPatriots<< tight end>> Rob Gronkowski had his bossâĢĻ
		Example 1: posted<|endoftext|>You should know this<< about>> offensive line coaches: they are large, demanding<< men>>
		Example 2: Media Day 2015ĊĊLSU<< defensive>> end Isaiah Washington (94) speaks<< to the>>
		Example 3:<< running backs>>," he said. .. Defensive<< end>> Carroll Phillips is improving and his injury is
		Example 4:<< line>>, with the left side âĢĶ namely<< tackle>> Byron Bell at<< tackle>> and<< guard>> Amini

		"[1,0,0,1,1]"

		Latent explanation: The word "guys" in the phrase "you guys".

		Test examples:

		Example 0: if you are<< comfortable>> with it. You<< guys>> support me in many other ways already and
		Example 1: birth control access<|endoftext|> but I assure you<< women>> in Kentucky aren't laughing as they struggle
		Example 2:âĢĻs gig! I hope you guys<< LOVE>> her, and<< please>> be nice,
		Example 3:American, told<< Hannity>> that âĢľyou<< guys>> are playing the race card.âĢĿ
		Example 4:<< the>><|endoftext|>ľI want to<< remind>> you all that 10 days ago (director Massimil

		"[0,0,0,0,0]"

		Latent explanation: "of" before words that start with a capital letter.

		Test examples:

		Example 0: climate, TomblinâĢĻs Chief<< of>> Staff Charlie Lorensen said.Ċ
		Example 1: no wonderworking relics, no true Body and Blood<< of>> Christ, no true Baptism
		Example 2:ĊĊDeborah Sathe, Head<< of>> Talent Development and Production at Film London,
		Example 3:ĊĊIt has been devised by Director<< of>> Public Prosecutions (DPP)
		Example 4: and fair investigation not even include the Director<< of>> Athletics? A· Finally, we believe the

		"[1,1,1,1,1]"
		"""

	\end{tcolorbox}

	\caption{Fuzzing score prompt (Llama 3.1 70B)}
	\label{fig:Llama_prompt_fuz}

\end{figure}

\end{document}